\documentclass{article}

\PassOptionsToPackage{numbers, sort&compress}{natbib}
\usepackage[eandd,preprint]{neurips_2026}

\usepackage[utf8]{inputenc} 
\usepackage[T1]{fontenc}    
\usepackage{hyperref}       
\usepackage{url}            
\usepackage{booktabs}       
\usepackage{amsfonts}       
\usepackage{nicefrac}       
\usepackage{microtype}      
\usepackage{xcolor}         
\usepackage{graphicx} 
\usepackage{amsmath}
\usepackage{rotating}
\usepackage[version=4]{mhchem}

\usepackage{xspace}
\usepackage{multiobjective}
\usepackage[colorinlistoftodos,prependcaption,textsize=tiny,disable]{todonotes} 

\newcommand{\ourroot}{\texttt{Planktonzilla}}
\newcommand{\ourthing}{\ourroot\xspace}
\newcommand{\ourdataset}{\ourroot\texttt{-17M}\xspace}

\title{\ourthing\\Multimodal dataset and models for understanding plankton ecosystems}

\author{%
  Alan Contreras, Luis Valenzuela, Luis Martí, and Nayat Sanchez-Pi\thanks{Dataset and models are publicly available at \url{https://huggingface.co/datasets/project-oceania/}. Code use to train and evaluate the models and generate the \ourdataset is available at \url{https://github.com/Inria-Chile/planktonzilla}.} \\
  Inria Chile Research Center\\
  Inria --- Institut national de recherche en sciences et technologies du numérique\\
  Av. Apoquindo 2827, Las Condes 7550268 Chile.\\
  \texttt{<name.lastname>@inria.cl}, \url{https://inria.cl}\\
}

\begin{document}

\maketitle

\todo[inline, color=orange!40]{ANONYMIZATION (r009): the author block above names four authors + Inria Chile in plaintext, but the package option is \texttt{[dandb]} (default double-blind). Pick a side before submission: (a) keep \texttt{[dandb]} and anonymize the author block; or (b) switch to \texttt{\textbackslash usepackage[preprint]\{neurips\_2025\}} for an explicitly single-blind submission (D\&B permits single-blind per the 2025 CFP).}

\begin{abstract}

Marine plankton underpin aquatic food webs and play a key role in global \ce{CO2} sequestration, making reliable species identification critical for understanding ocean health and climate feedbacks. Existing classification models perform well on individual collections but fail to generalize across instruments and environments due to isolated training datasets and inconsistent labels. To address this, we introduce \ourdataset, a unified dataset consolidating publicly available plankton image collections spanning thirteen imaging systems. It comprises 17.4 million images with standardized taxonomy and geo-environmental metadata, including 3.74 million plankton images spanning over 602 taxonomic classes, of which 201 are identified at the species level, making it the largest and most comprehensive plankton image dataset to date. Using this large-scale dataset, we perform a controlled comparison between supervised and CLIP-style image–text training on a shared ViT backbone. We find that a supervised classifier matches or exceeds CLIP-style training when trained using taxonomic lineage as text. We further observe that BioCLIP and BioCLIP2 perform poorly on plankton in zero-shot and few-shot settings. Leveraging \ourdataset improves plankton classification performance, highlighting the limitations of current biological foundation models in marine imaging domains.
\end{abstract}

\section{Introduction}
Plankton play a fundamental role in aquatic ecosystems, contributing roughly half of global oxygen production and mediating key biogeochemical cycles \citep{Kareinen2025}. Monitoring their diversity and abundance is therefore critical for understanding ocean health and anticipating environmental change. The advent of automated in‑situ imaging systems has led to an explosion of plankton image data, yet fine‑grained species identification remains challenging: acquisition devices, sampling regions, imaging conditions, and subtle morphological variation induce strong shifts across collections \citep{Kareinen2025}. As a result, models trained on a single source often perform poorly when applied to new datasets or previously unseen plankton taxa \citep{Kareinen2025}.

Although deep learning and self-supervised methods have improved plankton classification, most existing studies still rely on isolated datasets with limited taxonomic breadth \citep{Ciranni2025}. Moreover, labels across sources are inconsistently formatted and annotated at heterogeneous taxonomic ranks, preventing direct integration and making evaluation difficult. Consequently, there is no unified benchmark that supports reproducible comparisons across data sources, taxonomic levels, and model training paradigms.


We present \ourdataset, a multimodal dataset designed to address these limitations. By aggregating 17.4 million images from thirteen different imaging systems and pairing them with available georeferenced environmental metadata, we provide, to our knowledge, the most comprehensive coverage of plankton taxa to date. We standardize taxonomic annotations across all constituent datasets using the World Register of Marine Species (WoRMS), retaining each label at its deepest valid rank. Because only a small fraction of plankton samples are identified at species level (about 6\%) and approximately 30\% are annotated up to genus, our harmonization preserves coarser but valid annotations rather than forcing them to species. This process resolves naming discrepancies and enables consistent evaluation across sources.

Beyond dataset construction, \ourdataset is designed as an evaluation resource. We define reproducible training, validation, and test splits stratified by source dataset and taxonomic labels, together with an external unseen-class evaluation setting over held-out source datasets. We use this benchmark to compare supervised ViT classifiers with CLIP-style image--text models trained from taxonomic lineages, and to assess the transfer of biological foundation models such as BioCLIP and BioCLIP2. As a result, \ourdataset provides both a large-scale corpus and a practical benchmark for studying plankton recognition, taxonomic granularity, domain adaptation, and future cross-instrument robustness \citep{Irisson2026,Batrakhanov2024}. A detailed survey of related work and the challenges addressed by our dataset is provided in the next section.


%


\paragraph{Contributions.} Concretely, this paper contributes:
\begin{enumerate}
    \item \textbf{\ourdataset}, a multimodal plankton image benchmark aggregating 17.4 million images from 13 imaging systems, including 3.74 million plankton observations spanning 604 taxonomic classes, paired with available geo-environmental metadata---to our knowledge the largest harmonized plankton image corpus to date.
    
    \item A \textbf{WoRMS-grounded taxonomic harmonization} of all constituent datasets to the deepest valid rank, with explicit per-rank coverage statistics that make the limited species-level annotation transparent: only about 6\% of plankton samples reach species level, while roughly 30\% are annotated up to genus.
    
    \item A \textbf{per-source biodiversity characterization} based on Shannon diversity, Shannon evenness, novelty score, and taxonomic coverage, identifying which source datasets contribute unique taxonomic information and which provide complementary observations across instruments and regions.
    
    \item An \textbf{empirical benchmark on \ourdataset} comparing supervised ViT classifiers with CLIP-style image--text models trained from taxonomic lineages, and evaluating the transfer of biological foundation models such as BioCLIP and BioCLIP2. Our results show that models trained on \ourdataset improve plankton recognition relative to existing biological foundation models, including in zero-shot and few-shot settings over unseen plankton classes.
\end{enumerate}

\subsection{Related work}

Automated plankton image classification has traditionally followed a two-step pipeline: extracting handcrafted descriptors of size, shape, texture, or morphology, and then training conventional classifiers such as Random Forests or Support Vector Machines \cite{grosjean2004, gorsky2010}. The field has since shifted toward deep learning, particularly Convolutional Neural Networks (CNNs), which learn representations directly from raw plankton images and substantially improve performance on individual datasets \cite{luo2018, panaiotis2026, sofia_inria}. More recent self-supervised and transfer-learning approaches reduce the need for extensive annotation, but most evaluations remain tied to single collections.

This single-source evaluation paradigm limits the conclusions that can be drawn from existing benchmarks. Plankton images are strongly affected by the imaging instrument, optical configuration, sampling protocol, geographic region, and local community composition. Consequently, models trained on one dataset often fail under distribution shift, particularly when applied to new sources or unseen plankton categories \cite{moreno2012, gonzalez2017}. Recent multi-instrument and domain-adaptation benchmarks have started to address this issue \cite{Batrakhanov2024}, while large-scale comparisons show that performance depends not only on classifier complexity but also on data quality and consistency \cite{panaiotis2026}. However, public plankton collections remain fragmented and taxonomically heterogeneous. 

A second limitation is label inconsistency. Across datasets, semantically equivalent organisms may appear under different spellings, synonyms, naming conventions, or taxonomic resolutions. Some annotations reach species level, whereas others stop at genus, family, order, or coarser ranks. In \ourdataset, we address this by harmonizing all constituent datasets with the World Register of Marine Species (WoRMS), retaining each sample at its deepest valid taxonomic rank rather than forcing all labels to species.

A central comparison point is the BioCLIP family of biological foundation models. BioCLIP~\cite{bioclip} uses a CLIP-style image--text architecture trained on organism images paired with taxonomic lineages from TreeOfLife-10M. BioCLIP\,2~\cite{bioclip2} scales this approach to TreeOfLife-200M, reporting improved performance across broad biological recognition tasks. Yet plankton remains challenging: on the \emph{Plankton} benchmark in~\cite{bioclip2}---a 4,080-image split from WHOI-Plankton~\cite{whoi_dataset}---zero-shot accuracy remains low ($1.0\%$--$6.1\%$), and BioCLIP\,2 underperforms BioCLIP ($3.9\%$ vs.\ $6.1\%$).

Our work complements these efforts by providing a large-scale, standardized benchmark for evaluating plankton classifiers across heterogeneous public sources. We show that using this dataset improves plankton classification performance, and that both supervised and CLIP-style training help reduce the poor predictions observed in current foundation models, including BioCLIP-style approaches.


\section{Constructing the Dataset}

\subsection{Data sources}

\ourdataset aggregates publicly available plankton image datasets collected with heterogeneous imaging systems and sampling protocols. These sources differ substantially in acquisition device, geographic coverage, annotation format, and taxonomic resolution, making direct comparison and joint model training difficult without prior harmonization.

Because multiple datasets contained semantically equivalent labels with heterogeneous spelling and syntax, the dataset was standardized and enriched through the addition of auxiliary attributes. We incorporated taxonomy, proposed label, plankton indicator, root class indicator, a qualifier and geo-environmental metadata. A detailed description of all dataset features and columns is provided in Appendix~\ref{sce:appx-dataset}.

Taxonomic lineages were retrieved by querying the World Register of Marine Species (WoRMS) webservice \cite{WoRMS20260504} using the original annotations.\footnote{For EcoTaxa datasets, the last valid taxonomic rank was extracted from the provided lineage.} We extracted ranks from Kingdom to Species whenever available. When a valid lineage was found, the proposed label corresponded to the deepest valid taxonomic rank; otherwise, the original annotation was retained, which only occurred for non-plankton samples. This preserves the original taxonomic resolution rather than forcing all labels to species level.

Plankton presence and organism viability were determined using metadata from EcoTaxa datasets, while remaining samples were manually curated by experts through taxonomic assessment and visual inspection. This allowed us to distinguish plankton observations from detritus, inert particles, organism fragments, and imaging artifacts.

Geo-environmental variables (longitude, latitude, minimum depth, maximum depth, temperature, and salinity) were collected when publicly accessible via dataset webservice.\footnote{\url{https://ecotaxa.obs-vlfr.fr/api/docs}}\footnote{\url{https://ifcb-data.whoi.edu/timeline?dataset=mvco}} Metadata availability varied across sources, particularly among EcoTaxa, IFCB, and JEDI datasets.

\begin{table}[tb]
\caption{Characteristics of the datasets used for constructing \ourdataset.}\label{tbl:planktozilla-sources}
\centering
\begin{tabular}{lccc}
\toprule
\textbf{Dataset}         & \textbf{Instrument}          & \textbf{\# of samples}  & \textbf{\# of classes} \\ 
\midrule
GlobalUVP5Net \cite{globaluvp5}      & UVP5                & 7.41M          & 254  \\
WHOI-plankton \cite{whoi_dataset}      & IFCB                & 3.56M          & 103  \\
JEDI System / OCEANS\textunderscore CPICS \cite{jedidataset}    & CPICS                    & 1.92M          & 95   \\
ZooScanNet \cite{zooscannet}           & ZooScan             & 1.45M          & 120  \\
ZooCAMNet \cite{zoocamnet}             & ZooCAM              & 1.29M          & 93   \\
UVP6Net   \cite{uvp6net}               & UVP6                & 634K           & 54   \\
ISIISNet   \cite{isiisnet}             & ISIIS               & 408K           & 32   \\
PlanktoScope \cite{planktoscope}      & PlanktonScope                & 180K          & 263  \\
FlowCAMNet \cite{flowcamnet}           & FlowCAM             & 141K           & 38   \\
MedPlanktonSet \cite{medplanktonset}   & IFCB                & 77.3K          & 139  \\
SykeIFCB2022 \cite{sykeifcb2022}      & IFCB                & 63.1K          & 50  \\
PlanktonSet 1.0 \cite{planktonset1.0}      & ISIIS-2                & 60.7K          & 121  \\
SykeZoosCan2024 \cite{sykezooscan2024} & ZooScan             & 22.8K          & 20   \\
Zoolake  \cite{zoolake}                & DSPC                & 17.9K          & 35   \\
Lensless \cite{lensless}               & Lensless Microscope & 6.4K           & 10   \\ 
\bottomrule
\end{tabular}
\end{table}

After harmonization and enrichment, the resulting dataset contains 17.4M images. Among them, 3.74M correspond to plankton observations, covering 602 distinct classes, of which 201 are identified at the species level. The remaining samples correspond to non-plankton observations, including detritus (e.g., organism fragments or molts), inert material (e.g., sand or mineral particles), and imaging artifacts (e.g., blur, bubbles, or poor focus). These samples are retained to preserve realistic variability and support the development and evaluation of out-of-distribution (OOD) detection methods.

\begin{figure}[tb]
\centering
\includegraphics[width=\linewidth]{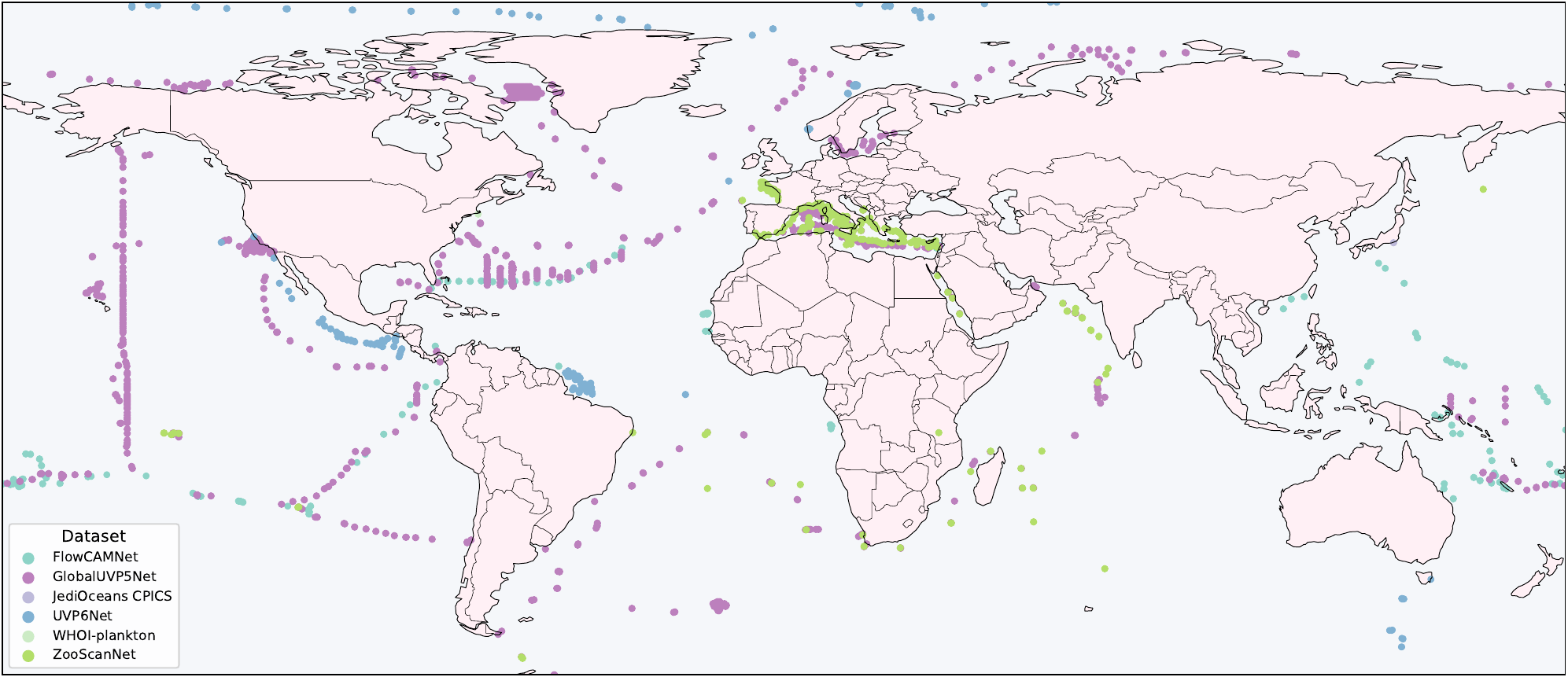}
\caption{Geographical distribution of samples composing \ourdataset identifying their corresponding source dataset. Only constituent datasets with publicly available geographic metadata are shown.}
\label{fig:geo_plot}
\end{figure}

By integrating datasets acquired with multiple instruments across different geographic regions, \ourdataset achieves broader spatial and taxonomic coverage than any individual constituent dataset. This expanded spatial representation is illustrated in Figure~\ref{fig:geo_plot}.

\subsection{Taxonomy description}

\ourdataset spans multiple taxonomic ranks for plankton, including 5 kingdoms, 29 phyla, 62 classes, 155 orders, 242 families, 289 genera, and 201 unique species. The complete taxonomic hierarchy represented in the dataset is illustrated in Figure~\ref{fig:tax_tree}.

To quantify the biodiversity contribution of each constituent dataset across taxonomic ranks, we computed Shannon diversity, Shannon evenness, novelty score, and taxonomic coverage. These metrics characterize both biodiversity and benchmark composition, identifying which sources contribute taxonomic richness, balanced class distributions, or complementary coverage.

The Shannon diversity index is defined as:

\begin{equation}
H' = - \sum_{i=1}^{S} p_i \log p_i
\end{equation}

where $S$ denotes the total number of classes at a given taxonomic rank, and $p_i$ represents the relative frequency of class $i$.

This metric captures both class richness and the uniformity of the class distribution. However, the Shannon index may assign higher values to highly uniform yet taxonomically limited distributions compared to distributions that are more taxonomically diverse but strongly imbalanced.

To isolate the uniformity component, we computed the Shannon evenness
\begin{equation}
E_{H} = \frac{H'}{\log S}
\end{equation}
where $\log S$ corresponds to the maximum possible Shannon diversity under a perfectly uniform class distribution. The evenness metric ranges from 0 to 1, with values close to 1 indicating highly balanced class distributions.

To further quantify the taxonomic contribution of each constituent dataset, we define a novelty score that measures the proportion of globally unique classes contributed by a given dataset. Let $C_d$ denote the set of classes present in dataset $d$, and let $C_{\text{global}}$ denote the set of all classes in \ourdataset. We define the set of unique contributions of dataset $d$ as
\begin{equation}
C_d^{\text{unique}} = \{ c \in C_d \;|\; c \notin C_{-d} \}
\end{equation}
where $C_{-d}$ represents the union of classes from all other datasets excluding $d$.

The novelty score is then defined as
\begin{equation}
N_d = \frac{|C_d^{\text{unique}}|}{|C_{\text{global}}|}
\end{equation}
where $|C_d^{\text{unique}}|$ denotes the number of classes exclusively contributed by dataset $d$, and $|C_{\text{global}}|$ is the total number of classes in the aggregated dataset.

Finally, because not all samples contain a complete taxonomic annotation, we compute the taxonomic coverage of each dataset across taxonomic ranks. This metric quantifies the proportion of samples in dataset $d$ that include a valid taxonomic assignment at the considered rank. The coverage is defined as
\begin{equation}
C_d = \frac{N_d^{\text{tax}}}{N_d}
\end{equation}
where $N_d^{\text{tax}}$ denotes the number of samples in dataset $d$ with a valid taxonomic annotation at the corresponding rank, and $N_d$ is the total number of samples in the dataset.

\begin{figure}[tb]
\centering
\includegraphics[width=\linewidth]{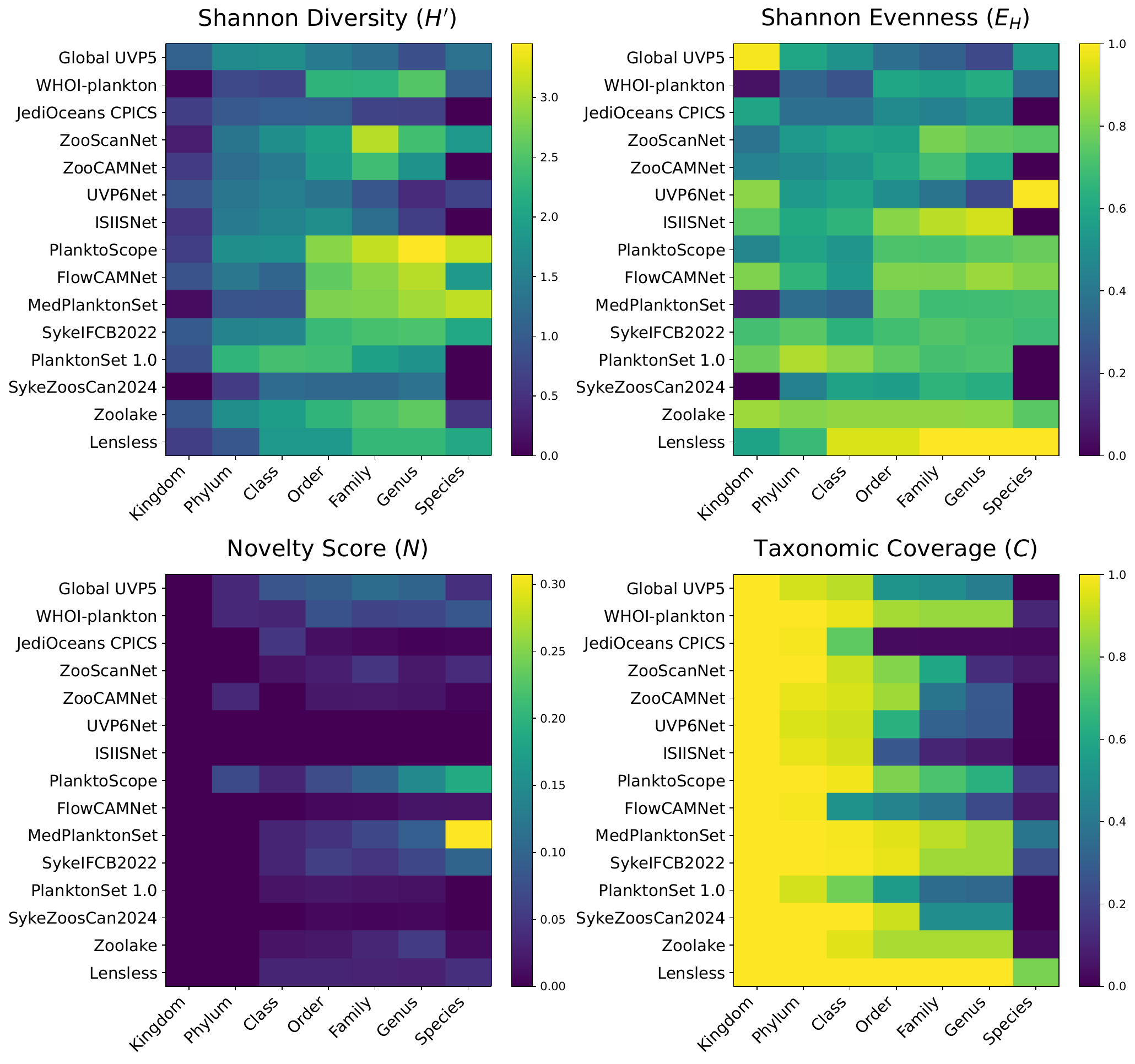}
\caption{Taxonomic diversity metrics computed for each constituent dataset of \ourdataset across taxonomic ranks. The figure reports the Shannon diversity index, Shannon evenness, novelty score, and taxonomic coverage.}
\label{fig:dataset_metrics}
\end{figure}

All these metrics are summarized in Figure~\ref{fig:dataset_metrics}. From this figure, it can be observed that PlanktoScope, FlowCAMNet, Lensless, MedPlanktonSet, ZooLake, and ZooScanNet are among the most diverse datasets across all taxonomic ranks. In addition to their high diversity, their taxonomic class distributions are relatively balanced, as reflected by their evenness values.

In contrast, ISIISNet, Jedi System/Oceans\textunderscore CPICS, SYKE ZooScan2024, PlanktonSet 1.0 and ZooCAMNet exhibit little to no species-level diversity. This is further confirmed by their taxonomic coverage, which is zero at the species rank.

When analyzing how informative each dataset is with respect to the taxonomic composition of \ourdataset, MedPlanktonSet stands out as one of the most informative datasets at the species level, contributing nearly 30\% of species that are not present in the other datasets. It also contributes unique classes at other taxonomic ranks. Similar contributions can be observed for PlanktoScope, WHOI-plankton, ZooScanNet, and FlowCAMNet, although to a lesser extent.

On the other hand, datasets such as ISIISNet, SykeZooScan2024, and UVP6Net contribute very few novel classes to the aggregated dataset. Nevertheless, they still provide complementary information by contributing additional observations of already existing classes obtained from different imaging instruments.

Finally, we observe that the constituent datasets exhibit relatively low species-level coverage, indicating that a large fraction of samples lack a complete taxonomic annotation. In fact, in \ourdataset only about 6\% of plankton samples contain a full species-level annotation, while approximately 30\% are annotated up to the genus level. Therefore, evaluations should be interpreted with respect to the deepest valid taxonomic rank available, rather than assuming complete species-level supervision.

\section{\ourthing Empirical Studies}

\subsection{Experimental Setup}

We evaluate \ourdataset through two complementary experimental settings designed to assess (i) in-domain classification and (ii) generalization to unseen plankton classes. Note that for all training and evaluation procedures described below, we exclusively use the plankton subset of \ourdataset, which consists of 3.74M images.

\paragraph{Experiment 1: In-domain evaluation.}
From the plankton subset, we first exclude four datasets (GlobalUVP5, PlanktoScope, PlanktonSet1.0, and SYKE-IFCB-2022), reserving them for the external evaluation setting. The remaining data is then partitioned into training (60\%), validation (20\%), and testing (20\%) sets. To preserve the original distributions, this division is stratified by both the source datasets and the taxonomic labels.

In this setting, we compare CLIP-style training against fully supervised classification using the same visual backbones (ViT-B/16 and ViT-L/14). CLIP models are trained using taxonomic lineage as text, following \cite{bioclip, bioclip2}, while supervised models are trained using a cross-entropy loss.

Since both training and test samples originate from the same underlying datasets, zero-shot and few-shot evaluation protocols are not applicable in this experiment. Instead, models are evaluated using standard classification metrics. This experiment isolates the effect of the training paradigm (CLIP vs supervised) under controlled conditions.

\paragraph{Experiment 2: Out-of-domain generalization.}
To evaluate generalization, we use the four held-out datasets (GlobalUVP5, PlanktoScope, PlanktonSet1.0, and SYKE-IFCB-2022) as a test set. From these, we select 220 plankton classes (113,089 samples) that are absent from the training data, ensuring a strict separation between seen and unseen categories.

This setting enables zero-shot and few-shot evaluation. We restrict this experiment to CLIP-based models to allow a fair comparison with BioCLIP and BioCLIP2, which are also evaluated under zero-shot and few-shot regimes. Few-shot classification is performed using the SimpleShot method \cite{simpleshot} on the visual embeddings. To account for variance in support set sampling, all one-shot and five-shot results are reported as the average Macro-F1 score across five random seeds.

This experiment provides a larger and more ecologically diverse evaluation than the plankton benchmark used in BioCLIP and BioCLIP2, which contains approximately 4,000 images. In contrast, our unseen-class evaluation uses 113,089 samples spanning 220 plankton classes across four held-out source datasets. This setting tests whether training on \ourdataset improves CLIP-based recognition of plankton categories not observed during training.

\paragraph{Training details.}
All models are trained exclusively on the 3.74M plankton samples from our defined training split. All CLIP models are trained using the OpenCLIP framework \cite{openclip}. We use ViT-B/16 initialized from OpenAI \cite{openai} and BioCLIP \cite{bioclip}, and ViT-L/14 initialized from LAION-2B \cite{laion2b} and BioCLIP2 \cite{bioclip2}. CLIP models are trained for 100 epochs using 64 H100 GPUs with a global batch size of 16,384. Supervised classifiers are trained for 20 epochs with a batch size of 64.

Model selection is based on the highest Macro-F1 score on the validation set. Additional hyperparameter configurations are provided in Appendix~\ref{sce:appx-training}.

\paragraph{Evaluation protocol.}
Performance is measured using Macro-F1 at each taxonomic rank. Labels are truncated at each level (kingdom to species), and samples are evaluated only at ranks for which a valid annotation is available. This setup assesses how well the models separate classes at increasing levels of taxonomic granularity: as evaluation moves from higher to lower ranks, performance reflects the ability to distinguish progressively finer taxa within the same parent groups.

For CLIP-based models, zero-shot predictions are obtained via similarity between image and text embeddings. For supervised models, predictions are obtained via softmax over class logits.

\begin{table}[tb]
\centering
\caption{Taxonomic classification performance measured using Macro-F1 score across different classification settings. Results are reported per taxonomic rank (Kingdom to Species) for ViT-B/16 and ViT-L/14 backbones, with pre-training sources indicated in parentheses. Zero-shot and few-shot results are reported for BioCLIP and BioCLIP2 models, while standard classification corresponds to models trained on \ourdataset. For all few-shot evaluations, the reported values represent the average score across five random seeds.}
\label{tab:metrics}
\small
\setlength{\tabcolsep}{3pt}

\resizebox{\linewidth}{!}{
\begin{tabular}{lccccccc}
\toprule
\multicolumn{1}{c}{\textbf{Model}} &
\textbf{Kingdom} &
\textbf{Phylum} &
\textbf{Class} &
\textbf{Order} &
\textbf{Family} &
\textbf{Genus} &
\textbf{Species} \\
\midrule

\textit{Zero-Shot Classification} & & & & & & & \\ \midrule
ViT-B/16 {\scriptsize (BioCLIP)} & 0.184 & 0.042 & 0.022 & 0.010 & 0.007 & 0.005 & 0.004 \\
ViT-L/14 {\scriptsize (BioCLIP2)} & 0.287 & 0.086 & 0.046 & 0.034 & 0.020 & 0.016 & 0.013 \\ 

\midrule
\textit{One-Shot Classification (SimpleShot)} & & & & & & & \\ \midrule
ViT-B/16 {\scriptsize (BioCLIP)} & 0.411 & 0.244 & 0.198 & 0.170 & 0.141 & 0.134 & 0.128 \\
ViT-L/14 {\scriptsize (BioCLIP2)} & 0.457 & 0.284 & 0.229 & 0.193 & 0.160 & 0.148 & 0.137 \\ 

\midrule
\textit{Five-Shot Classification (SimpleShot)} & & & & & & & \\ \midrule
ViT-B/16 {\scriptsize (BioCLIP)} & 0.496 & 0.361 & 0.298 & 0.270 & 0.235 & 0.224 & 0.215 \\
ViT-L/14 {\scriptsize (BioCLIP2)} & 0.527 & 0.399 & 0.333 & 0.296 & 0.254 & 0.239 & 0.224 \\

\midrule
\textit{Standard Classification} & & & & & & & \\ \midrule
ViT-B/16 {\scriptsize (BioCLIP + \ourdataset)} & 0.932 & 0.858 & 0.810 & 0.800 & 0.768 & 0.752 & 0.734 \\
ViT-B/16 {\scriptsize (OpenAI + \ourdataset)} & 0.961 & 0.905 & 0.865 & 0.853 & 0.822 & 0.797 & 0.786 \\ 
\quad $\hookrightarrow$ Supervised classifier & \textbf{0.979} & \textbf{0.954} & \textbf{0.911} & \textbf{0.905} & \textbf{0.882} & \textbf{0.878} & \textbf{0.867} \\ \addlinespace[2ex]

ViT-L/14 {\scriptsize (BioCLIP2 + \ourdataset)} & 0.964 & 0.888 & 0.856 & 0.852 & 0.833 & 0.823 & 0.816 \\
ViT-L/14 {\scriptsize (LAION-2B + \ourdataset)} & 0.967 & 0.927 & 0.878 & 0.873 & 0.845 & 0.839 & 0.825 \\ 
\quad $\hookrightarrow$ Supervised classifier & \textbf{0.980} & \textbf{0.958} & \textbf{0.911} & \textbf{0.906} & \textbf{0.889} & \textbf{0.882} & \textbf{0.874} \\

\bottomrule
\end{tabular}
}
\end{table}

\begin{table}[tb]
\centering
\caption{Average taxonomic classification performance (Macro-F1 score) evaluated on unseen plankton classes across test datasets. Results are reported per taxonomic rank (Kingdom to Species) for ViT-B/16 and ViT-L/14 backbones, with pre-training sources indicated in parentheses. Zero-shot and few-shot results are reported for BioCLIP, BioCLIP2, and models trained on the plankton subset of \ourdataset. Best results per backbone size are highlighted in bold. Few-shot results (one-shot and five-shot) represent the average performance across five random seeds.}
\label{tab:metrics_unseen_classes}
\small
\setlength{\tabcolsep}{3pt}

\resizebox{\linewidth}{!}{
\begin{tabular}{lccccccc}
\toprule
\multicolumn{1}{c}{\textbf{Model}} &
\textbf{Kingdom} &
\textbf{Phylum} &
\textbf{Class} &
\textbf{Order} &
\textbf{Family} &
\textbf{Genus} &
\textbf{Species} \\
\midrule

\textit{Zero-Shot Classification} & & & & & & & \\ \midrule
ViT-B/16 {\scriptsize (BioCLIP)} & 0.346 & 0.102 & 0.032 & 0.018 & 0.013 & 0.011 & 0.010 \\
ViT-B/16 {\scriptsize (BioCLIP + \ourdataset)} & 0.282 & 0.100 & 0.057 & 0.039 & 0.030 & 0.026 & 0.023 \\
ViT-B/16 {\scriptsize (OpenAI + \ourdataset)} & \textbf{0.408} & \textbf{0.204} & \textbf{0.118} & \textbf{0.088} & \textbf{0.072} & \textbf{0.065} & \textbf{0.055} \\ \addlinespace[1.5ex]

ViT-L/14 {\scriptsize (BioCLIP2)} & 0.292 & 0.114 & 0.052 & 0.036 & 0.026 & 0.019 & 0.015 \\ 
ViT-L/14 {\scriptsize (LAION-2B + \ourdataset)} & \textbf{0.428} & \textbf{0.238} & \textbf{0.132} & 0.098 & \textbf{0.089} & 0.079 & 0.068 \\ 
ViT-L/14 {\scriptsize (BioCLIP2 + \ourdataset)} & 0.391 & 0.206 & 0.123 & \textbf{0.100} & \textbf{0.089} & \textbf{0.080} & \textbf{0.070} \\

\midrule
\textit{One-Shot Classification (SimpleShot)} & & & & & & & \\ \midrule
ViT-B/16 {\scriptsize (BioCLIP)} & 0.444 & 0.227 & 0.162 & 0.133 & 0.121 & 0.113 & 0.102 \\
ViT-B/16 {\scriptsize (BioCLIP + \ourdataset)} & 0.510 & 0.246 & 0.171 & 0.137 & 0.121 & 0.111 & 0.101 \\
ViT-B/16 {\scriptsize (OpenAI + \ourdataset)} & \textbf{0.549} & \textbf{0.296} & \textbf{0.204} & \textbf{0.166} & \textbf{0.150} & \textbf{0.139} & \textbf{0.128} \\ \addlinespace[1.5ex]

ViT-L/14 {\scriptsize (BioCLIP2)} & 0.509 & 0.274 & 0.188 & 0.153 & 0.140 & 0.127 & 0.117 \\
ViT-L/14 {\scriptsize (LAION-2B + \ourdataset)} & \textbf{0.536} & 0.293 & 0.196 & 0.165 & 0.151 & 0.140 & 0.127 \\ 
ViT-L/14 {\scriptsize (BioCLIP2 + \ourdataset)} & 0.535 & \textbf{0.319} & \textbf{0.220} & \textbf{0.177} & \textbf{0.164} & \textbf{0.156} & \textbf{0.142} \\ 

\midrule
\textit{Five-Shot Classification (SimpleShot)} & & & & & & & \\ \midrule
ViT-B/16 {\scriptsize (BioCLIP)} & 0.552 & 0.326 & 0.252 & 0.226 & 0.209 & 0.194 & 0.180 \\
ViT-B/16 {\scriptsize (BioCLIP + \ourdataset)} & 0.607 & 0.328 & 0.246 & 0.218 & 0.199 & 0.187 & 0.172 \\
ViT-B/16 {\scriptsize (OpenAI + \ourdataset)} & \textbf{0.673} & \textbf{0.398} & \textbf{0.291} & \textbf{0.259} & \textbf{0.241} & \textbf{0.228} & \textbf{0.213} \\ \addlinespace[1.5ex]

ViT-L/14 {\scriptsize (BioCLIP2)} & 0.629 & 0.374 & 0.290 & 0.252 & 0.235 & 0.217 & 0.202 \\
ViT-L/14 {\scriptsize (LAION-2B + \ourdataset)} & \textbf{0.664} & 0.384 & 0.278 & 0.257 & 0.242 & 0.230 & 0.214 \\ 
ViT-L/14 {\scriptsize (BioCLIP2 + \ourdataset)} & 0.646 & \textbf{0.411} & \textbf{0.295} & \textbf{0.269} & \textbf{0.256} & \textbf{0.244} & \textbf{0.226} \\
\bottomrule
\end{tabular}
}
\end{table}

\subsection{Results and Analysis}

\paragraph{In-domain performance.}
Table~\ref{tab:metrics} reports classification performance on \ourdataset using the stratified test split described above. Models trained on \ourdataset significantly outperform BioCLIP and BioCLIP2 across all taxonomic ranks, confirming that general-purpose biological foundation models have limited direct transfer to plankton recognition.

Among all approaches, fully supervised classifiers achieve the best performance. Despite not using explicit taxonomic information during training, these models consistently outperform CLIP-based models across all ranks. This suggests that strong visual supervision alone is sufficient to learn representations that align with taxonomic structure. This behavior is further illustrated in Figure~\ref{fig:taxo_tsne}, where learned embeddings naturally organize according to taxonomy.

CLIP-based models trained on \ourdataset also achieve strong performance, but remain below their supervised counterparts. This indicates that incorporating taxonomy as text does not necessarily improve classification performance in a fully supervised setting.

\paragraph{Generalization to unseen plankton classes.}
Table~\ref{tab:metrics_unseen_classes} reports performance on 220 plankton classes not seen during training, evaluated across four independent datasets.

In the zero-shot setting, BioCLIP and BioCLIP2 show low performance across all taxonomic ranks. CLIP-based models trained on \ourdataset generally achieve higher scores, with consistent but moderate improvements across most ranks.

These trends persist in the few-shot regime. For both one-shot and five-shot classification, models trained on \ourdataset tend to outperform BioCLIP and BioCLIP2 across both backbone sizes. While the margins are not large, the improvements are consistent across taxonomic levels, including finer-grained ranks such as genus and species.

Overall, these results indicate that training with \ourdataset leads to more reliable predictions on unseen plankton classes and helps mitigate the performance limitations observed in existing biological foundation models.

\section{Conclusion}

We introduced \ourdataset, a multimodal dataset aggregating 17.4 million images with broad plankton coverage, standardized taxonomy, and available geo-environmental metadata. Our experiments focus on a 3.74M image plankton subset. Using this subset, we observe that general-purpose biological foundation models, such as BioCLIP and BioCLIP2, show limited zero-shot transfer to plankton taxa.

Our results provide two main observations regarding model training paradigms. First, for in-domain classification, a fully supervised ViT classifier performs comparably to, or slightly better than, CLIP-style models trained with taxonomic lineage as text. This suggests that, in this specific setting, visual supervision is highly effective for learning representations aligned with plankton taxonomy. Second, in out-of-domain scenarios, CLIP models trained on the plankton subset improve zero-shot and few-shot recognition of plankton classes absent from training when compared with existing biological foundation models.

A secondary intended impact of \ourdataset is to support more reproducible evaluation and pretraining of plankton classifiers, with potential use in operational ocean-observatory pipelines, plankton-based biogeochemistry, and climate-relevant ecosystem monitoring. We see two plausible paths to negative impact, in particular:
\begin{itemize}
    \item Geo-environmental metadata in the dataset --- particularly fine-resolution coordinates of historical sampling sites --- could in principle be combined with abundance estimates to inform unauthorized fishing or conservation-arbitrage decisions; we mitigate this by releasing only the metadata that the source datasets had already publicly distributed and by encouraging downstream users to follow the Ocean Biodiversity Information System (OBIS) data-use norms.\footnote{See \url{https://manual.obis.org/policy.html}.}
    \item Classifier outputs may inherit and amplify the geographic and taxonomic biases described above; we discourage uncritical use of \ourdataset-pretrained models for community-composition estimates outside the regions and instruments well represented in the corpus.
\end{itemize}

We release the complete \ourdataset to support the community in expanding taxonomic coverage and developing models for marine ecosystems.


\newpage

\bibliographystyle{apalike}
\bibliography{references}


\newpage

\appendix

\section{Additional Details of \ourdataset Dataset}\label{sce:appx-dataset}

Each sample in \ourdataset contains a set of features describing its visual content, taxonomic annotation, provenance, and geo-environmental context. Table~\ref{tab:dataset_features} provides a detailed description of all features.

\begin{table*}[h]
\centering
\small
\begin{tabular}{lll}
\toprule
\textbf{Feature} & \textbf{Type} & \textbf{Description} \\
\midrule
image & image & PIL Image object in RGB format with variable resolution \\
dataset & string & Source dataset identifier (e.g., flowcamnet, isiisnet, whoi-plankton) \\
original\_label & string & Class label from the original dataset \\
original\_path & string & Original file path in source dataset \\
Kingdom & string & Taxonomic Kingdom classification \\
Phylum & string & Taxonomic Phylum classification \\
Class & string & Taxonomic Class classification \\
Order & string & Taxonomic Order classification \\
Family & string & Taxonomic Family classification \\
Genus & string & Taxonomic Genus classification \\
Species & string & Taxonomic Species classification \\
proposed\_label & string & Harmonized label across datasets \\
plankton & bool & Boolean indicating if the object is plankton \\
root\_class & string & High-level category (e.g., living, detritus, inert, artifact) \\
qualifier & string & Additional qualifier (e.g., full\_body, part, larvae, egg) \\
Latitude & float32 & Geographic latitude of the observation (degrees) \\
Longitude & float32 & Geographic longitude of the observation (degrees) \\
Depth\_min & float32 & Minimum sampling depth in meters \\
Depth\_max & float32 & Maximum sampling depth in meters \\
Temperature & float32 & Water temperature at sampling location \\
Humidity & float32 & Humidity of the sample \\
ObjID & string & Unique object identifier from the source dataset (Ecotaxa or IFCB) \\
\bottomrule
\end{tabular}
\caption{Overview of the features available for each sample in \ourdataset.}
\label{tab:dataset_features}
\end{table*}

\begin{figure}
\centering
\includegraphics[width=\linewidth]{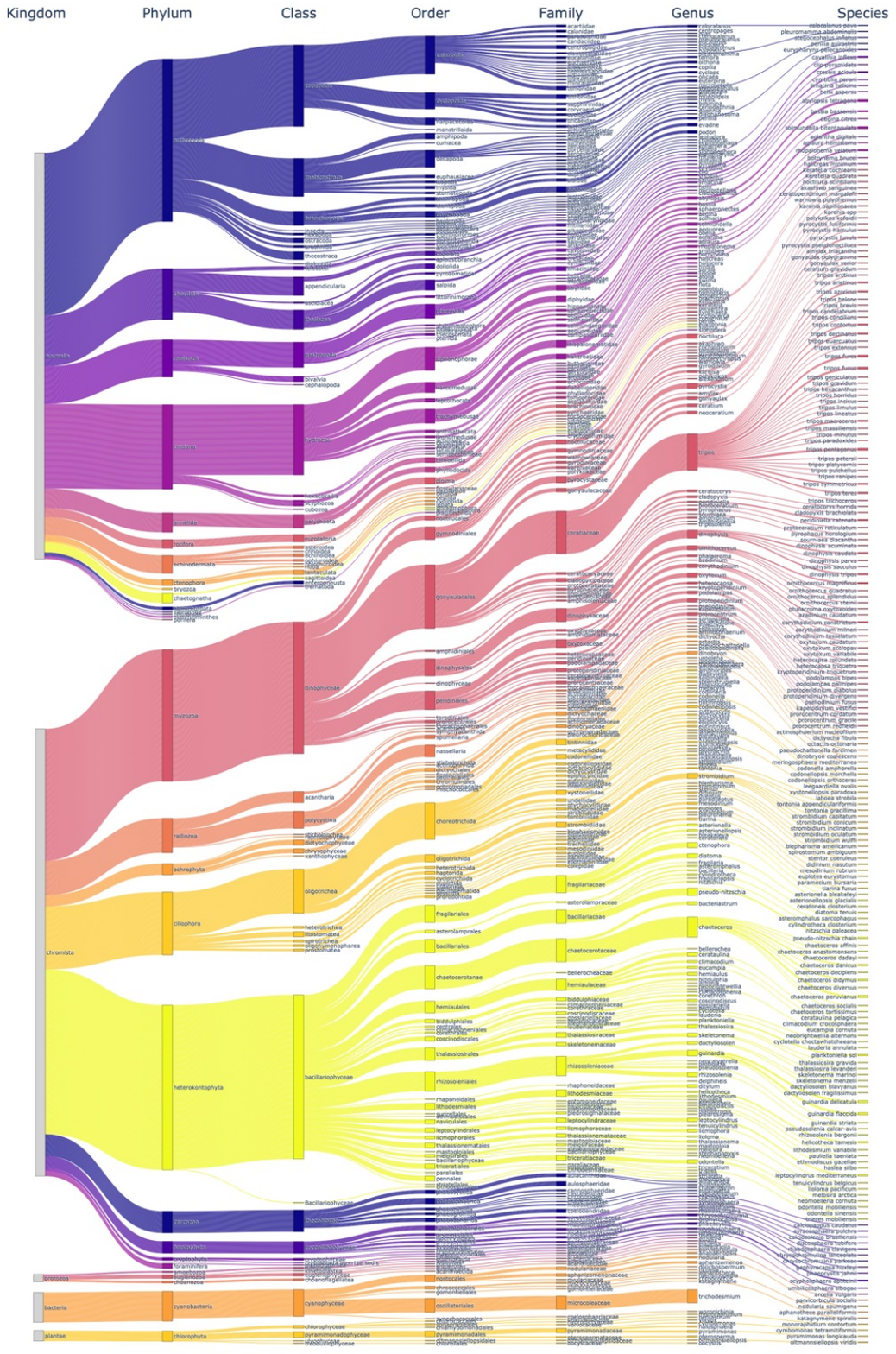}
\caption{Taxonomy tree of the plankton present in \ourdataset.}
\label{fig:tax_tree}
\end{figure}

Figure \ref{fig:tax_tree} describes the taxonomy of the plankton present in the dataset.

\newpage

\section{Training details}\label{sce:appx-training}

Tables~\ref{tab:clip_hparams} and~\ref{tab:sup_hparams} summarize the hyperparameters used for CLIP-based and supervised models, respectively. The

\begin{table}[tb]
\centering
\small

\begin{minipage}{0.48\linewidth}
\centering
\caption{Training hyperparameters for CLIP-based models.}
\label{tab:clip_hparams}
\setlength{\tabcolsep}{5pt}
\begin{tabular}{lcc}
\toprule
\textbf{Hyperparameter} & \textbf{ViT-B/16} & \textbf{ViT-L/14} \\
\midrule
Input size & $224 \times 224$ & $224 \times 224$ \\
Batch size & 16,384 & 16,384 \\
Learning rate & $1 \times 10^{-4}$ & $1 \times 10^{-4}$ \\
Warm-up steps & 1,000 & 1,000 \\
Scheduler & Cosine decay & Cosine decay \\
Max epochs & 100 & 100 \\
Weight decay & 0.2 & 0.2 \\
\bottomrule
\end{tabular}
\end{minipage}
\hfill
\begin{minipage}{0.48\linewidth}
\centering
\caption{Training hyperparameters for supervised classifiers.}
\label{tab:sup_hparams}
\setlength{\tabcolsep}{5pt}
\begin{tabular}{lcc}
\toprule
\textbf{Hyperparameter} & \textbf{ViT-B/16} & \textbf{ViT-L/14} \\
\midrule
Input size & $224 \times 224$ & $224 \times 224$ \\
Batch size & 64 & 64 \\
Learning rate & $1 \times 10^{-4}$ & $1 \times 10^{-4}$ \\
Warm-up steps & 1000 & 1000 \\
Scheduler & Cosine decay & Cosine decay \\
Max epochs & 20 & 20 \\
Weight decay & 0.2 & 0.2 \\
\bottomrule
\end{tabular}
\end{minipage}

\end{table}

Generating the full Planktonzilla dataset from scratch required approximately three continuous weeks of processing on a single node with 48 CPU cores and 512~GB of RAM. Regarding model training, individual supervised baseline runs took 2 to 3.5 hours, while training the CLIP-based models required 10 to 15 hours per run. In total, accounting for all configurations, the overall training process took approximately 55 hours using $64\times$ NVIDIA H100 GPUs. All computational workloads were executed on a net-zero carbon footprint HPC facility that operates entirely independent of fossil fuel energy.

\section{Aditionals experiments}\label{sce:additional_experiments}

Table~\ref{tab:metrics_unseen_datasets} details the average taxonomic classification performance evaluated on the full hold-out test set (GlobalUVP5Net, PlanktoScope, PlanktonSet 1.0, and SykeIFCB2022). This comprehensive evaluation is performed over a total of 821,211 unseen plankton samples.

\begin{table}[tb]
\centering
\caption{Average taxonomic classification performance (Macro-F1 score) evaluated on unseen plankton test datasets. Results are reported per taxonomic rank (Kingdom to Species) for ViT-B/16 and ViT-L/14 backbones, with pre-training sources indicated in parentheses. Zero-shot and few-shot results are reported for BioCLIP, BioCLIP2, and models trained on the plankton subset of \ourdataset. Best results per backbone size are highlighted in bold. Few-shot results (one-shot and five-shot) represent the average performance across five random seeds.}
\label{tab:metrics_unseen_datasets}
\small
\setlength{\tabcolsep}{3pt}

\resizebox{\linewidth}{!}{
\begin{tabular}{lccccccc}
\toprule
\multicolumn{1}{c}{\textbf{Model}} &
\textbf{Kingdom} &
\textbf{Phylum} &
\textbf{Class} &
\textbf{Order} &
\textbf{Family} &
\textbf{Genus} &
\textbf{Species} \\
\midrule

\textit{Zero-Shot Classification} & & & & & & & \\ \midrule
ViT-B/16 {\scriptsize (BioCLIP)} & 0.346 & 0.102 & 0.032 & 0.018 & 0.013 & 0.011 & 0.010 \\
ViT-B/16 {\scriptsize (BioCLIP + \ourdataset)} & 0.282 & 0.100 & 0.057 & 0.039 & 0.030 & 0.026 & 0.023 \\
ViT-B/16 {\scriptsize (OpenAI + \ourdataset)} & \textbf{0.408} & \textbf{0.204} & \textbf{0.118} & \textbf{0.088} & \textbf{0.072} & \textbf{0.065} & \textbf{0.055} \\ \addlinespace[1.5ex]

ViT-L/14 {\scriptsize (BioCLIP2)} & 0.292 & 0.114 & 0.052 & 0.036 & 0.026 & 0.019 & 0.015 \\ 
ViT-L/14 {\scriptsize (LAION-2B + \ourdataset)} & \textbf{0.428} & \textbf{0.238} & \textbf{0.132} & 0.098 & \textbf{0.089} & 0.079 & 0.068 \\ 
ViT-L/14 {\scriptsize (BioCLIP2 + \ourdataset)} & 0.391 & 0.206 & 0.123 & \textbf{0.100} & \textbf{0.089} & \textbf{0.080} & \textbf{0.070} \\

\midrule
\textit{One-Shot Classification (SimpleShot)} & & & & & & & \\ \midrule
ViT-B/16 {\scriptsize (BioCLIP)} & 0.444 & 0.227 & 0.162 & 0.133 & 0.121 & 0.113 & 0.102 \\
ViT-B/16 {\scriptsize (BioCLIP + \ourdataset)} & 0.510 & 0.246 & 0.171 & 0.137 & 0.121 & 0.111 & 0.101 \\
ViT-B/16 {\scriptsize (OpenAI + \ourdataset)} & \textbf{0.549} & \textbf{0.296} & \textbf{0.204} & \textbf{0.166} & \textbf{0.150} & \textbf{0.139} & \textbf{0.128} \\ \addlinespace[1.5ex]

ViT-L/14 {\scriptsize (BioCLIP2)} & 0.509 & 0.274 & 0.188 & 0.153 & 0.140 & 0.127 & 0.117 \\
ViT-L/14 {\scriptsize (LAION-2B + \ourdataset)} & \textbf{0.536} & 0.293 & 0.196 & 0.165 & 0.151 & 0.140 & 0.127 \\ 
ViT-L/14 {\scriptsize (BioCLIP2 + \ourdataset)} & 0.535 & \textbf{0.319} & \textbf{0.220} & \textbf{0.177} & \textbf{0.164} & \textbf{0.156} & \textbf{0.142} \\ 

\midrule
\textit{Five-Shot Classification (SimpleShot)} & & & & & & & \\ \midrule
ViT-B/16 {\scriptsize (BioCLIP)} & 0.552 & 0.326 & 0.252 & 0.226 & 0.209 & 0.194 & 0.180 \\
ViT-B/16 {\scriptsize (BioCLIP + \ourdataset)} & 0.607 & 0.328 & 0.246 & 0.218 & 0.199 & 0.187 & 0.172 \\
ViT-B/16 {\scriptsize (OpenAI + \ourdataset)} & \textbf{0.673} & \textbf{0.398} & \textbf{0.291} & \textbf{0.259} & \textbf{0.241} & \textbf{0.228} & \textbf{0.213} \\ \addlinespace[1.5ex]

ViT-L/14 {\scriptsize (BioCLIP2)} & 0.629 & 0.374 & 0.290 & 0.252 & 0.235 & 0.217 & 0.202 \\
ViT-L/14 {\scriptsize (LAION-2B + \ourdataset)} & \textbf{0.664} & 0.384 & 0.278 & 0.257 & 0.242 & 0.230 & 0.214 \\ 
ViT-L/14 {\scriptsize (BioCLIP2 + \ourdataset)} & 0.646 & \textbf{0.411} & \textbf{0.295} & \textbf{0.269} & \textbf{0.256} & \textbf{0.244} & \textbf{0.226} \\

\bottomrule
\end{tabular}
}
\end{table}



\begin{figure}[tb]
\centering
\includegraphics[width=\linewidth]{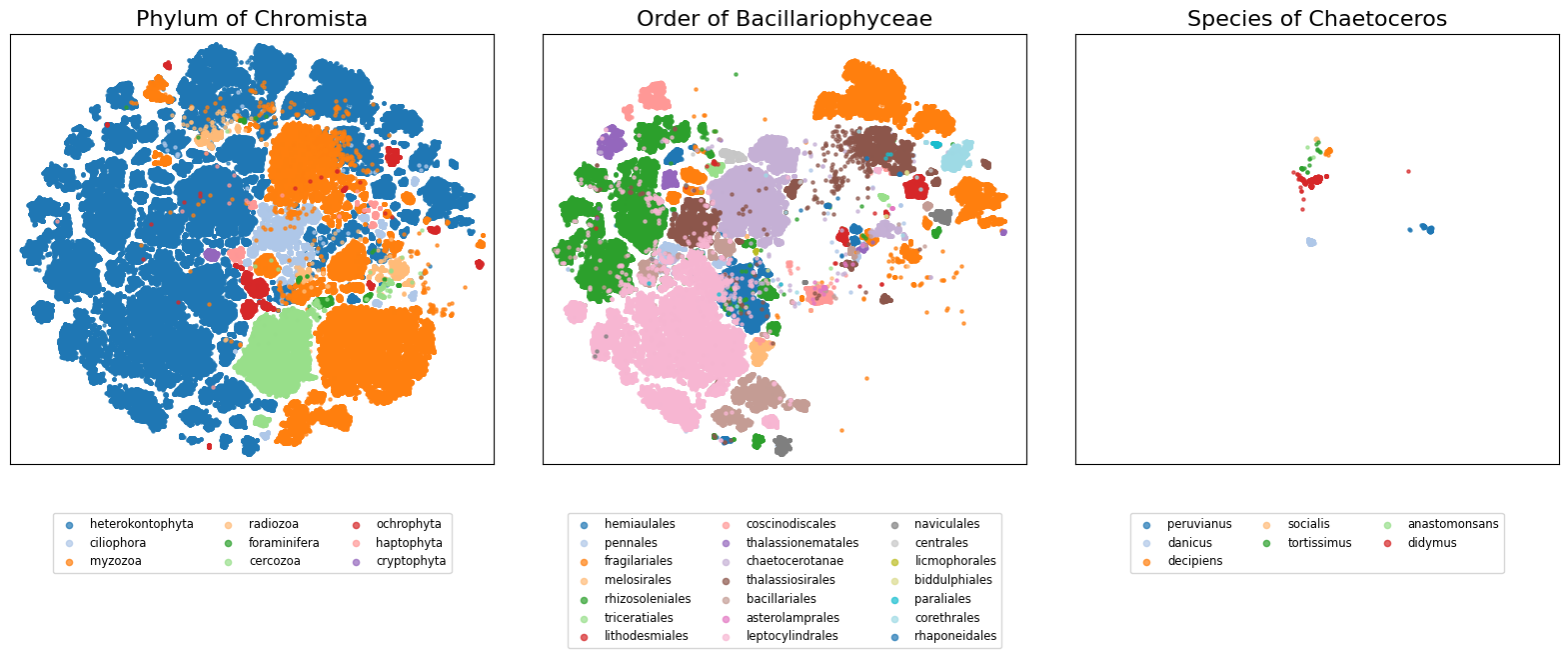}
\caption{t-SNE visualization of the visual features extracted by the ViT-B/16 classifier across different taxonomic ranks within the Kingdom \emph{Chromista}. The left panel shows the different phyla, the center panel displays the different orders within the phylum \emph{Bacillariophyceae}, and the right panel presents the various species within the genus \emph{Chaetoceros}.}
\label{fig:taxo_tsne}
\end{figure}

Furthermore, as a qualitative visual inspection to demonstrate that supervised models can implicitly learn taxonomic hierarchies relying solely on visual features, Figure~\ref{fig:taxo_tsne} presents a specific case study focusing on the Kingdom \emph{Chromista}.

\section{Data, Code, and Licensing Availability}
\label{appx:licenses}

\begin{itemize}
    \item The complete \ourdataset dataset, including the 17.4M images and models, is publicly available on HuggingFace at \url{https://huggingface.co/datasets/project-oceania/}.
    \item All code required to train and evaluate the models and preprocess and generate the \ourdataset dataset is available on GitHub at \url{https://github.com/Inria-Chile/planktonzilla}.
    \item Our aggregated dataset is released under the CC-BY-NC-4.0 license. Detailed licensing information, URLs, and attribution for each source dataset are provided in Table \ref{tab:source_datasets_licenses}.
\end{itemize}

Table \ref{tab:source_datasets_licenses} details the 15 individual source datasets that comprise \ourdataset, including their respective imaging systems, data volume, taxonomic classes, and licensing information. We provide direct URLs to the original repositories to ensure proper attribution.

\begin{table}[tb]
\centering
\caption{Overview of the source datasets included in \ourdataset, detailing their respective instruments, number of images and classes, licensing, and original hosting platforms.}
\label{tab:source_datasets_licenses}
\begin{tabular}{@{}l@{}c@{}c@{}c@{}c@{}c@{}}
\toprule
\textbf{Dataset} & \textbf{Instrument} & \textbf{Images} & \textbf{Classes} & \textbf{License} & \textbf{Source} \\ 
\midrule
GlobalUVP5Net & UVP5 & 7.41M & 254 & CC-BY-NC-4.0 & \href{https://www.seanoe.org/data/00964/107583/}{SEANOE} \\
WHOI-plankton & IFCB & 3.56M & 103 & MIT & \href{https://github.com/hsosik/WHOI-Plankton}{GitHub} \\
JEDI System / OCEANS\textunderscore CPICS & CPICS & 1.92M & 95 & CC-BY-SA-4.0 & \href{https://dbarchive.biosciencedbc.jp/data/jedisystem-oceansdb/LATEST/README_e.html#Sec3}{JAMSTEC} \\
ZooScanNet & ZooScan & 1.45M & 120 & CC-BY-NC-4.0 & \href{https://www.seanoe.org/data/00446/55741/}{SEANOE} \\
ZooCAMNet & ZooCAM & 1.29M & 93 & CC-BY-NC-4.0 & \href{https://www.seanoe.org/data/00907/101928/}{SEANOE} \\
UVP6Net & UVP6 & 634K & 54 & CC-BY-NC-4.0 & \href{https://www.seanoe.org/data/00908/101948/}{SEANOE} \\
ISIISNet & ISIIS & 408K & 32 & CC-BY-NC-4.0 & \href{https://www.seanoe.org/data/00908/101950/}{SEANOE} \\
PlanktoScope & PlanktonScope & 180K & 263 & CC-BY-NC-4.0 & \href{https://www.seanoe.org/data/00989/110078/}{SEANOE} \\
FlowCAMNet & FlowCAM & 141K & 38 & CC-BY-NC-4.0 & \href{https://www.seanoe.org/data/00908/101961/}{SEANOE} \\
MedPlanktonSet & IFCB & 77.3K & 139 & CC-BY-NC-4.0 & \href{https://zenodo.org/records/15471023}{Zenodo} \\
SykeIFCB2022 & IFCB & 63.1K & 50 & CC-BY-NC-4.0 & \href{https://b2share.eudat.eu/records/abf913e5a6ad47e6baa273ae0ed6617a}{B2SHARE} \\
PlanktonSet 1.0 & ISIIS-2 & 60.7K & 121 & CC-BY-NC-4.0 & \href{https://www.ncei.noaa.gov/archive/archive-management-system/OAS/bin/prd/jquery/accession/download/127422}{NOAA NCEI} \\
SykeZoosCan2024 & ZooScan & 22.8K & 20 & CC-BY-NC-4.0 & \href{https://etsin.fairdata.fi/dataset/6fa42787-9772-41a5-a6fc-0dde489ed908}{FairData} \\
Zoolake & DSPC & 17.9K & 35 & CC-BY-4.0 & \href{https://opendata.eawag.ch/dataset/52b6ba86-5ecb-448c-8c01-eec7cb209dc7}{EAWAG OpenData} \\
Lensless &  Microscope & 6.4K & 10 & CC-BY-4.0 & \href{https://ibm.ent.box.com/v/PlanktonData}{IBM Box} \\
\bottomrule
\end{tabular}
\end{table}


\newpage
\section*{NeurIPS Paper Checklist}

\begin{enumerate}

\item {\bf Claims}
    \item[] Question: Do the main claims made in the abstract and introduction accurately reflect the paper's contributions and scope?
    \item[] Answer: \answerYes{} 
    \item[] Justification: The abstract and introduction state the main contributions: the construction of Planktonzilla-17M, a large-scale standardized plankton dataset, and a controlled comparison between supervised and CLIP-style training. They also claim that training on this dataset helps mitigate the poor performance of existing biological foundation models on plankton. These claims are supported by the dataset description (Section 2) and experimental results (Section 3), which show improved performance in both in-domain and zero/few-shot settings.
    \item[] Guidelines:
    \begin{itemize}
        \item The answer \answerNA{} means that the abstract and introduction do not include the claims made in the paper.
        \item The abstract and/or introduction should clearly state the claims made, including the contributions made in the paper and important assumptions and limitations. A \answerNo{} or \answerNA{} answer to this question will not be perceived well by the reviewers. 
        \item The claims made should match theoretical and experimental results, and reflect how much the results can be expected to generalize to other settings. 
        \item It is fine to include aspirational goals as motivation as long as it is clear that these goals are not attained by the paper. 
    \end{itemize}

\item {\bf Limitations}
    \item[] Question: Does the paper discuss the limitations of the work performed by the authors?
    \item[] Answer: \answerYes{} 
    \item[] Justification: Sections 1, 2.2, and 4 discuss limitations, including low taxonomic resolution at the species level and the risk of inheriting or amplifying geographic and taxonomic biases.
    \item[] Guidelines:
    \begin{itemize}
        \item The answer \answerNA{} means that the paper has no limitation while the answer \answerNo{} means that the paper has limitations, but those are not discussed in the paper. 
        \item The authors are encouraged to create a separate ``Limitations'' section in their paper.
        \item The paper should point out any strong assumptions and how robust the results are to violations of these assumptions (e.g., independence assumptions, noiseless settings, model well-specification, asymptotic approximations only holding locally). The authors should reflect on how these assumptions might be violated in practice and what the implications would be.
        \item The authors should reflect on the scope of the claims made, e.g., if the approach was only tested on a few datasets or with a few runs. In general, empirical results often depend on implicit assumptions, which should be articulated.
        \item The authors should reflect on the factors that influence the performance of the approach. For example, a facial recognition algorithm may perform poorly when image resolution is low or images are taken in low lighting. Or a speech-to-text system might not be used reliably to provide closed captions for online lectures because it fails to handle technical jargon.
        \item The authors should discuss the computational efficiency of the proposed algorithms and how they scale with dataset size.
        \item If applicable, the authors should discuss possible limitations of their approach to address problems of privacy and fairness.
        \item While the authors might fear that complete honesty about limitations might be used by reviewers as grounds for rejection, a worse outcome might be that reviewers discover limitations that aren't acknowledged in the paper. The authors should use their best judgment and recognize that individual actions in favor of transparency play an important role in developing norms that preserve the integrity of the community. Reviewers will be specifically instructed to not penalize honesty concerning limitations.
    \end{itemize}

\item {\bf Theory assumptions and proofs}
    \item[] Question: For each theoretical result, does the paper provide the full set of assumptions and a complete (and correct) proof?
    \item[] Answer: \answerNA{} 
    \item[] Justification: The paper focuses on the introduction of a new dataset and empirical benchmarking. It does not present theoretical results.
    \item[] Guidelines:
    \begin{itemize}
        \item The answer \answerNA{} means that the paper does not include theoretical results. 
        \item All the theorems, formulas, and proofs in the paper should be numbered and cross-referenced.
        \item All assumptions should be clearly stated or referenced in the statement of any theorems.
        \item The proofs can either appear in the main paper or the supplemental material, but if they appear in the supplemental material, the authors are encouraged to provide a short proof sketch to provide intuition. 
        \item Inversely, any informal proof provided in the core of the paper should be complemented by formal proofs provided in appendix or supplemental material.
        \item Theorems and Lemmas that the proof relies upon should be properly referenced. 
    \end{itemize}

    \item {\bf Experimental result reproducibility}
    \item[] Question: Does the paper fully disclose all the information needed to reproduce the main experimental results of the paper to the extent that it affects the main claims and/or conclusions of the paper (regardless of whether the code and data are provided or not)?
    \item[] Answer: \answerYes{} 
    \item[] Justification: Experimental setups are detailed in Section 3.1 and Appendix B. All code and data are publicly available (see Dataset Availability section).
    \item[] Guidelines:
    \begin{itemize}
        \item The answer \answerNA{} means that the paper does not include experiments.
        \item If the paper includes experiments, a \answerNo{} answer to this question will not be perceived well by the reviewers: Making the paper reproducible is important, regardless of whether the code and data are provided or not.
        \item If the contribution is a dataset and\slash or model, the authors should describe the steps taken to make their results reproducible or verifiable. 
        \item Depending on the contribution, reproducibility can be accomplished in various ways. For example, if the contribution is a novel architecture, describing the architecture fully might suffice, or if the contribution is a specific model and empirical evaluation, it may be necessary to either make it possible for others to replicate the model with the same dataset, or provide access to the model. In general. releasing code and data is often one good way to accomplish this, but reproducibility can also be provided via detailed instructions for how to replicate the results, access to a hosted model (e.g., in the case of a large language model), releasing of a model checkpoint, or other means that are appropriate to the research performed.
        \item While NeurIPS does not require releasing code, the conference does require all submissions to provide some reasonable avenue for reproducibility, which may depend on the nature of the contribution. For example
        \begin{enumerate}
            \item If the contribution is primarily a new algorithm, the paper should make it clear how to reproduce that algorithm.
            \item If the contribution is primarily a new model architecture, the paper should describe the architecture clearly and fully.
            \item If the contribution is a new model (e.g., a large language model), then there should either be a way to access this model for reproducing the results or a way to reproduce the model (e.g., with an open-source dataset or instructions for how to construct the dataset).
            \item We recognize that reproducibility may be tricky in some cases, in which case authors are welcome to describe the particular way they provide for reproducibility. In the case of closed-source models, it may be that access to the model is limited in some way (e.g., to registered users), but it should be possible for other researchers to have some path to reproducing or verifying the results.
        \end{enumerate}
    \end{itemize}

\item {\bf Open access to data and code}
    \item[] Question: Does the paper provide open access to the data and code, with sufficient instructions to faithfully reproduce the main experimental results, as described in supplemental material?
    \item[] Answer: \answerYes{} 
    \item[] Justification: Our dataset and code are publicly available at \url{https://huggingface.co/datasets/project-oceania/planktonzilla-17M} and \url{https://github.com/Inria-Chile/planktonzilla}.
    \item[] Guidelines:
    \begin{itemize}
        \item The answer \answerNA{} means that paper does not include experiments requiring code.
        \item Please see the NeurIPS code and data submission guidelines (\url{https://neurips.cc/public/guides/CodeSubmissionPolicy}) for more details.
        \item While we encourage the release of code and data, we understand that this might not be possible, so \answerNo{} is an acceptable answer. Papers cannot be rejected simply for not including code, unless this is central to the contribution (e.g., for a new open-source benchmark).
        \item The instructions should contain the exact command and environment needed to run to reproduce the results. See the NeurIPS code and data submission guidelines (\url{https://neurips.cc/public/guides/CodeSubmissionPolicy}) for more details.
        \item The authors should provide instructions on data access and preparation, including how to access the raw data, preprocessed data, intermediate data, and generated data, etc.
        \item The authors should provide scripts to reproduce all experimental results for the new proposed method and baselines. If only a subset of experiments are reproducible, they should state which ones are omitted from the script and why.
        \item At submission time, to preserve anonymity, the authors should release anonymized versions (if applicable).
        \item Providing as much information as possible in supplemental material (appended to the paper) is recommended, but including URLs to data and code is permitted.
    \end{itemize}

\item {\bf Experimental setting/details}
    \item[] Question: Does the paper specify all the training and test details (e.g., data splits, hyperparameters, how they were chosen, type of optimizer) necessary to understand the results?
    \item[] Answer: \answerYes{} 
    \item[] Justification: Training and test details are specified in Section 3.1, and hyperparameters are documented in Appendix B.
    \item[] Guidelines:
    \begin{itemize}
        \item The answer \answerNA{} means that the paper does not include experiments.
        \item The experimental setting should be presented in the core of the paper to a level of detail that is necessary to appreciate the results and make sense of them.
        \item The full details can be provided either with the code, in appendix, or as supplemental material.
    \end{itemize}

\item {\bf Experiment statistical significance}
    \item[] Question: Does the paper report error bars suitably and correctly defined or other appropriate information about the statistical significance of the experiments?
    \item[] Answer: \answerYes{} 
    \item[] Justification: For different tasks, different random seeds were used to obtain the few-shot Macro-F1 metrics.
    \item[] Guidelines:
    \begin{itemize}
        \item The answer \answerNA{} means that the paper does not include experiments.
        \item The authors should answer \answerYes{} if the results are accompanied by error bars, confidence intervals, or statistical significance tests, at least for the experiments that support the main claims of the paper.
        \item The factors of variability that the error bars are capturing should be clearly stated (for example, train/test split, initialization, random drawing of some parameter, or overall run with given experimental conditions).
        \item The method for calculating the error bars should be explained (closed form formula, call to a library function, bootstrap, etc.)
        \item The assumptions made should be given (e.g., Normally distributed errors).
        \item It should be clear whether the error bar is the standard deviation or the standard error of the mean.
        \item It is OK to report 1-sigma error bars, but one should state it. The authors should preferably report a 2-sigma error bar than state that they have a 96\% CI, if the hypothesis of Normality of errors is not verified.
        \item For asymmetric distributions, the authors should be careful not to show in tables or figures symmetric error bars that would yield results that are out of range (e.g., negative error rates).
        \item If error bars are reported in tables or plots, the authors should explain in the text how they were calculated and reference the corresponding figures or tables in the text.
    \end{itemize}

\item {\bf Experiments compute resources}
    \item[] Question: For each experiment, does the paper provide sufficient information on the computer resources (type of compute workers, memory, time of execution) needed to reproduce the experiments?
    \item[] Answer: \answerYes{} 
    \item[] Justification: Section 4 and Appendix B specify the computational hardware used and provide estimates for the GPU-hours required for training.
    \item[] Guidelines:
    \begin{itemize}
        \item The answer \answerNA{} means that the paper does not include experiments.
        \item The paper should indicate the type of compute workers CPU or GPU, internal cluster, or cloud provider, including relevant memory and storage.
        \item The paper should provide the amount of compute required for each of the individual experimental runs as well as estimate the total compute. 
        \item The paper should disclose whether the full research project required more compute than the experiments reported in the paper (e.g., preliminary or failed experiments that didn't make it into the paper). 
    \end{itemize}
    
\item {\bf Code of ethics}
    \item[] Question: Does the research conducted in the paper conform, in every respect, with the NeurIPS Code of Ethics \url{https://neurips.cc/public/EthicsGuidelines}?
    \item[] Answer: \answerYes{} 
    \item[] Justification: The research adheres to the NeurIPS Code of Ethics by properly attributing source datasets, respecting original licenses, and ensuring open access to the aggregated data.
    \item[] Guidelines:
    \begin{itemize}
        \item The answer \answerNA{} means that the authors have not reviewed the NeurIPS Code of Ethics.
        \item If the authors answer \answerNo, they should explain the special circumstances that require a deviation from the Code of Ethics.
        \item The authors should make sure to preserve anonymity (e.g., if there is a special consideration due to laws or regulations in their jurisdiction).
    \end{itemize}

\item {\bf Broader impacts}
    \item[] Question: Does the paper discuss both potential positive societal impacts and negative societal impacts of the work performed?
    \item[] Answer: \answerYes{} 
    \item[] Justification: Section 4 discusses both the potential benefits for ocean observatories, biogeochemistry, and climate monitoring, and the risks associated with metadata misuse for unauthorized fishing and biased predictions in underrepresented regions.
    \item[] Guidelines:
    \begin{itemize}
        \item The answer \answerNA{} means that there is no societal impact of the work performed.
        \item If the authors answer \answerNA{} or \answerNo, they should explain why their work has no societal impact or why the paper does not address societal impact.
        \item Examples of negative societal impacts include potential malicious or unintended uses (e.g., disinformation, generating fake profiles, surveillance), fairness considerations (e.g., deployment of technologies that could make decisions that unfairly impact specific groups), privacy considerations, and security considerations.
        \item The conference expects that many papers will be foundational research and not tied to particular applications, let alone deployments. However, if there is a direct path to any negative applications, the authors should point it out. For example, it is legitimate to point out that an improvement in the quality of generative models could be used to generate Deepfakes for disinformation. On the other hand, it is not needed to point out that a generic algorithm for optimizing neural networks could enable people to train models that generate Deepfakes faster.
        \item The authors should consider possible harms that could arise when the technology is being used as intended and functioning correctly, harms that could arise when the technology is being used as intended but gives incorrect results, and harms following from (intentional or unintentional) misuse of the technology.
        \item If there are negative societal impacts, the authors could also discuss possible mitigation strategies (e.g., gated release of models, providing defenses in addition to attacks, mechanisms for monitoring misuse, mechanisms to monitor how a system learns from feedback over time, improving the efficiency and accessibility of ML).
    \end{itemize}
    
\item {\bf Safeguards}
    \item[] Question: Does the paper describe safeguards that have been put in place for responsible release of data or models that have a high risk for misuse (e.g., pre-trained language models, image generators, or scraped datasets)?
    \item[] Answer: \answerNA{} 
    \item[] Justification: The released resources do not carry a significant risk of malicious misuse.
    \item[] Guidelines:
    \begin{itemize}
        \item The answer \answerNA{} means that the paper poses no such risks.
        \item Released models that have a high risk for misuse or dual-use should be released with necessary safeguards to allow for controlled use of the model, for example by requiring that users adhere to usage guidelines or restrictions to access the model or implementing safety filters. 
        \item Datasets that have been scraped from the Internet could pose safety risks. The authors should describe how they avoided releasing unsafe images.
        \item We recognize that providing effective safeguards is challenging, and many papers do not require this, but we encourage authors to take this into account and make a best faith effort.
    \end{itemize}

\item {\bf Licenses for existing assets}
    \item[] Question: Are the creators or original owners of assets (e.g., code, data, models), used in the paper, properly credited and are the license and terms of use explicitly mentioned and properly respected?
    \item[] Answer: \answerYes{} 
    \item[] Justification: Existing assets and their licenses are documented in Table 8.
    \item[] Guidelines:
    \begin{itemize}
        \item The answer \answerNA{} means that the paper does not use existing assets.
        \item The authors should cite the original paper that produced the code package or dataset.
        \item The authors should state which version of the asset is used and, if possible, include a URL.
        \item The name of the license (e.g., CC-BY 4.0) should be included for each asset.
        \item For scraped data from a particular source (e.g., website), the copyright and terms of service of that source should be provided.
        \item If assets are released, the license, copyright information, and terms of use in the package should be provided. For popular datasets, \url{paperswithcode.com/datasets} has curated licenses for some datasets. Their licensing guide can help determine the license of a dataset.
        \item For existing datasets that are re-packaged, both the original license and the license of the derived asset (if it has changed) should be provided.
        \item If this information is not available online, the authors are encouraged to reach out to the asset's creators.
    \end{itemize}

\item {\bf New assets}
    \item[] Question: Are new assets introduced in the paper well documented and is the documentation provided alongside the assets?
    \item[] Answer: \answerYes{} 
    \item[] Justification: The new dataset is documented and released with proper licensing.
    \item[] Guidelines:
    \begin{itemize}
        \item The answer \answerNA{} means that the paper does not release new assets.
        \item Researchers should communicate the details of the dataset\slash code\slash model as part of their submissions via structured templates. This includes details about training, license, limitations, etc. 
        \item The paper should discuss whether and how consent was obtained from people whose asset is used.
        \item At submission time, remember to anonymize your assets (if applicable). You can either create an anonymized URL or include an anonymized zip file.
    \end{itemize}

\item {\bf Crowdsourcing and research with human subjects}
    \item[] Question: For crowdsourcing experiments and research with human subjects, does the paper include the full text of instructions given to participants and screenshots, if applicable, as well as details about compensation (if any)? 
    \item[] Answer: \answerNA{} 
    \item[] Justification: The research does not involve crowdsourcing or human subjects.
    \item[] Guidelines:
    \begin{itemize}
        \item The answer \answerNA{} means that the paper does not involve crowdsourcing nor research with human subjects.
        \item Including this information in the supplemental material is fine, but if the main contribution of the paper involves human subjects, then as much detail as possible should be included in the main paper. 
        \item According to the NeurIPS Code of Ethics, workers involved in data collection, curation, or other labor should be paid at least the minimum wage in the country of the data collector. 
    \end{itemize}

\item {\bf Institutional review board (IRB) approvals or equivalent for research with human subjects}
    \item[] Question: Does the paper describe potential risks incurred by study participants, whether such risks were disclosed to the subjects, and whether Institutional Review Board (IRB) approvals (or an equivalent approval/review based on the requirements of your country or institution) were obtained?
    \item[] Answer: \answerNA{} 
    \item[] Justification: The research does not involve human subjects.
    \item[] Guidelines:
    \begin{itemize}
        \item The answer \answerNA{} means that the paper does not involve crowdsourcing nor research with human subjects.
        \item Depending on the country in which research is conducted, IRB approval (or equivalent) may be required for any human subjects research. If you obtained IRB approval, you should clearly state this in the paper. 
        \item We recognize that the procedures for this may vary significantly between institutions and locations, and we expect authors to adhere to the NeurIPS Code of Ethics and the guidelines for their institution. 
        \item For initial submissions, do not include any information that would break anonymity (if applicable), such as the institution conducting the review.
    \end{itemize}

\item {\bf Declaration of LLM usage}
    \item[] Question: Does the paper describe the usage of LLMs if it is an important, original, or non-standard component of the core methods in this research? Note that if the LLM is used only for writing, editing, or formatting purposes and does \emph{not} impact the core methodology, scientific rigor, or originality of the research, declaration is not required.
    \item[] Answer: \answerNA{} 
    \item[] Justification: LLMs were not used as a core component in this research.
    \item[] Guidelines:
    \begin{itemize}
        \item The answer \answerNA{} means that the core method development in this research does not involve LLMs as any important, original, or non-standard components.
        \item Please refer to our LLM policy in the NeurIPS handbook for what should or should not be described.
    \end{itemize}

\end{enumerate}

\end{document}